\documentclass[a4paper, 12pt, oneside]{amsart}
\usepackage{graphicx}
\usepackage{covington}
\usepackage{CJKutf8}
\newcommand{\zh}[1]{\begin{CJK}{UTF8}{min}#1\end{CJK}}
\usepackage{eulervm}
\usepackage{amsmath, amsthm}
\usepackage{tikz-cd}
\usepackage{pgf, tikz}
\usetikzlibrary{arrows,decorations.markings,automata}
\PassOptionsToPackage{usenames,dvipsnames,svgnames}{xcolor}  
\DeclareMathOperator{\fv}{FVect}
\DeclareMathOperator{\pgrp}{PGrp}
\renewcommand{\vec}[1]{\mathbf{#1}}
\newcommand{\fsub}[1]{#1_{\phantom{r}}}
\newcommand{\fsup}[1]{#1^{\phantom{r}}}
\newcommand{\fss}[1]{#1^{\phantom{r}}_{\phantom{r}}}
\theoremstyle{definition}
\newtheorem{definition}{Definition}
\newtheorem{metarule}{Metarule}
\newtheorem{eg}{Example}
\usepackage{hyperref}
\hypersetup{
  colorlinks = true,
  citecolor = {purple!70!black},
  linkcolor = {purple!70!black},
  urlcolor = {purple!70!black}
}

\begin{document}

\title{An algebraic approach to translating japanese}
\author{Valentin Boboc}
\address{School of Mathematics, The University of Manchester, Alan Turing Building, Oxford Road, Manchester M13 9PL, United Kingdom}
\email{valentinboboc@icloud.com}

\begin{abstract}
We use Lambek's pregroups and the framework of compositional distributional models of language (``DisCoCat'') to study translations from Japanese to English as pairs of functors. Adding decorations to pregroups we show how to handle word order changes between languages.
\end{abstract}
\maketitle
\section{Introduction}\label{sec:intro}
Language is traditionally viewed as possessing both an empirical aspect -- one learns language by practising language -- and a compositional aspect -- the view that the meaning of a complex phrase is fully determined by its structure and the meanings of its constituent parts. 

In order to efficiently exploit the compositional nature of languages, a popular way of modelling natural languages is a categorical compositional distributional model, abbreviated ``DisCoCat'' (\cite{coecke2010mathematical}). Languages are modelled as functors from a category that interprets grammar (``compositional'') to a category that interprets semantics (``distributional''). 

The compositional part is responsible for evaluating whether phrases or sentences are well formed by calculating the overall grammatical type of a phrase from the grammatical types of its individual parts. There are several algebraic methods for modelling the grammar of a natural language. In the present article we choose the well-established model of pregroup grammars. Pregroups were introduced in \cite{lambek1997type} to replace the algebra of residuated monoids in order to model grammatical types, their juxtapositions, and reductions. Pregroup calculus has been applied to formally represent the syntax of several natural languages such as: French (\cite{bargelli2001french}), German (\cite{lambek2003german}), Persian (\cite{sadrzadeh2007persian}), Arabic (\cite{bargelli2001arabic}), Japanese (\cite{cardinal2002algebraic}), and Latin (\cite{casadio2005latin}). 

The distributional part assigns meanings to individual words by associating to them, for example, statistical co-occurrence vectors (\cite{mitchell2008vector}). The ``DisCoCat'' model is thus a way of interpreting compositions of meanings via grammatical structure. 

In this article we study the notion of translating between compositional distributional models of language by analysing translation from Japanese into English. On the compositional side, a translation is a strong monoidal functor. It is easy to demonstrate that such a functor is too rigid to handle the translation of even simple phrases between languages which have different word order. We show that one can keep using the gadget of monoidal functors as long as the underlying pregroup grammars are decorated with additional structure.

We begin by introducing basic notions about the compact closed categories we work with, namely pregroups and finitely generated vector spaces and define our notion of translation functor. Next, we give an introduction to basic Japanese grammar and the pregroup structure we use to model it. Finally, we introduce notions of pregroup decorations and use them to give a structured framework for translating Japanese sentences.

\section{Theoretical background}\label{sec:background}
\subsection{Compact closed structures}
The key to the ``DisCoCat'' model is that both the category of pregroups and the category of finitely generated vector spaces are \emph{compact closed categories}. This allows for compositional characteristics of grammar to be incorporated into the distributional spaces of meaning. 

For completeness, we provide here a definition of compact closure. The reader is encouraged to consult (\cite{kelly1980coherence}) for a more complete and technical reference.

\begin{definition}
A \emph{compact closed category} is a category $\mathcal{C}$ together with a bifunctor \[-\otimes - : \mathcal{C} \times \mathcal{C} \to \mathcal{C},\] called tensor product, which is associative up to natural isomorphism and possesses a two-sided identity element $I$, and each object $A \in \mathcal{C}$ has a right dual $A^r$ and a left dual $A^\ell$ with the following morphisms
\begin{align*}
    A \otimes A^r \xrightarrow{\varepsilon^r_A} I \xrightarrow{\eta^r_A} A^r \otimes A,\\
    A^\ell \otimes A \xrightarrow{\varepsilon^\ell_A} I \xrightarrow{\eta^\ell_A} A \otimes A^\ell.
\end{align*}
Moreover, the $\varepsilon$ and $\eta$ maps satisfy the ``yanking'' conditions:
\begin{align*}
    &(1_A \otimes \varepsilon^\ell_A) \circ (\eta^\ell_A \otimes 1_A) = 1_A &&(\varepsilon^r_A \otimes 1_A)\circ (1_A \otimes \eta^r_A) = 1_A\\
    &(\varepsilon^\ell_A \otimes 1_A)\circ (1_{A^\ell} \otimes \eta^\ell_A) = 1_{A^\ell} &&(1_{A^r} \otimes \varepsilon^r_A)\circ (\eta^r_A \otimes 1_{A^r}) = 1_{A^r}.
\end{align*}
\end{definition}

The upshot of compact closure is that we want to have elements which ``cancel each other out'' and we can decompose the identity into a product. 
\subsection{Recalling pregroups}
\begin{definition}
A \emph{pregroup} is a tuple $(P, \cdot, 1, -^\ell, -^r, \leq)$ where $(P, \cdot, 1, \leq)$ is a partially ordered monoid and the unary operations $-^\ell, -^r$ (the left and the right dual) satisfy for all $x \in P$ the following relations:
\begin{align*}
    &x \cdot x^r \leq 1
    &&x^\ell \cdot x \leq 1\\
    &1 \leq x^r \cdot x
    &&1 \leq x \cdot x^\ell.
\end{align*}
\end{definition}
The operation sign $\cdot$ is omitted unless it is relevant. It is immediate to check that the following relations hold in every pregroup:
\begin{align*}
    &1^\ell = 1 = 1^r
    &&(x^\ell)^r = x = (x^r)^\ell\\
    &(xy)^\ell = y^\ell x^\ell \text{ and } (xy)^r = y^rx^r
    &&\text{if } x\leq y \text{ then } y^\ell \leq x^\ell \text{ and } y^r \leq x^r. 
\end{align*}

We model the grammar of a natural language by freely generating a pregroup from a set of grammatical types. Each word in the dictionary is assigned an element of the pregroup which corresponds to its linguistic function, e.g. noun, verb, adjective, etc. A string of words is interpreted by multiplying the elements assigned to the constituent parts in syntactic order. If a string of words satisfies the relation $w_1w_2\ldots w_n \leq s$ we say that the string \emph{reduces} to the type $s$. 
\begin{eg} Suppose there are two grammatical types: noun $n$ and sentence $s$. Grammar is modelled as the free pregroup $\pgrp(\{n, s\})$. Consider the sentence ``Pigeons eat bread.'' We assign the type $n$ to ``pigeons'' and ``bread'' and the type $n^r s n^\ell$ to the transitive verb ``eat.'' The sentence overall has type $n(n^r s n^\ell) n$ and the following reductions hold:
\[n(n^r s n^\ell)n =(nn^r) s (n^\ell n) \leq (1) s (n^\ell n) \leq s (1) \leq s.\]
In this case we say that ``Pigeons eat bread'' is a well-formed sentence since in the pregroup $\pgrp(\{n,s\})$ the phrase reduces to the correct type. 
\end{eg}

The two individual reductions could have been performed in a different order. Lambek's \emph{Switching Lemma} \cite[Proposition 2]{lambek1997type} tells us that in any computation performed in a freely generated pregroup, we may assume without loss of generality that all contractions precede all expansions. 

A pregroup can be viewed as a compact closed category. The objects of the category are the elements of the pregroup. There is an arrow $x \to y$ if and only if $x \leq y$, and the tensor product is given by the pregroup operation: $x \otimes y = xy$. The morphisms $\varepsilon^r, \varepsilon^\ell, \eta^r, \eta^\ell$ are defined in the obvious way. In terms of the $\varepsilon$ and $\eta$ maps, the reductions in this example can be represented as:
\[(\varepsilon^\ell_n \otimes 1_s \otimes \varepsilon^\ell_n)(n\otimes (n^r \otimes s \otimes n^\ell)\otimes n) \to s.\]
\subsection{Meaning space} We encode the semantic structure of a natural language into the category of finitely generated vector spaces, which we denote by $\fv$. The arrows are linear transformations, and there is a natural monoidal structure given by the linear algebraic tensor product with unit $\mathbb{R}$, which also happens to be symmetric: $V \otimes W \simeq W \otimes V$. This implies that $V^\ell \simeq V^r \simeq V^*$, where the latter denotes the dual vector space. 

Fixing a basis $\{\vec{v}_i\}$ for the vector space $V$ we get moreover that $V \simeq V^*$ and the structure morphisms of compact closure are given by 
\begin{align*}
    &\varepsilon_V = \varepsilon^r_V = \varepsilon^\ell_V : V^* \otimes V \to \mathbb{R} \text{ where } \sum_{i,j} a_{ij} v_i \otimes v_j \mapsto \sum_{i,j} a_{ij} \langle v_i | v_j \rangle\\
    &\eta_V = \eta^r_V = \eta^\ell_V : \mathbb{R} \to V \otimes V^* \text{ where } 1 \mapsto \sum_{i} v_i \otimes v_i \text{ extended linearly}. 
\end{align*}

If we denote by $P$ both the pregroup and the corresponding category, the bridge between grammar and semantics is given by a strong monoidal functor 
\[F : P \to \fv,\]
which we call a \emph{functorial language model}. The functor assigns vector spaces to atomic types: $F(1) = I$, $F(n) = N$ (the vector space of nouns), $F(s) = S$ (the vector space of sentences), etc. For words in $P$, monoidality tells us that $F(x\otimes y) = F(x) \otimes F(y)$. The compact closure is also preserved: $F(x^\ell) = F(x^r) = F(x)^*$. For example, we can interpret the transitive verb ``eat'' with type $n^r s n^\ell$ as a vector in \[F(n^r \otimes s \otimes n^\ell) = F(n^r) \otimes F(s) \otimes F(n^\ell) = F(n)^* \otimes F(s) \otimes F(n)^* = N \otimes S \otimes N.\]
Pregroup reductions in $P$ can be interpreted as semantic reductions in $\fv$ using the corresponding $\varepsilon$ and $\eta$ maps. The reductions associated to a transitive verb are then given by
\[F(\varepsilon^r_n \otimes 1_s \otimes \varepsilon^\ell_n) = F(\varepsilon^r_n) \otimes F(1_s) \otimes F(\varepsilon^\ell_n) = F(\varepsilon_n)^* \otimes F(1_s) \otimes F(\varepsilon_n)^* = \varepsilon_N \otimes 1_S \otimes \varepsilon_N.\]
The meaning of a sentence or phrase is derived by interpreting the pregroup reduction as the correponding semantic reduction of the tensor product of distributional meanings of individual words in the phrase. The previous example ``Pigeons eat bread'' is interpreted as \[F(\varepsilon^r_n \otimes 1_s \otimes \varepsilon^\ell_n)(\vec{Pigeons}\otimes \vec{eat}\otimes \vec{bread}).\]
\subsection{Translating between functorial language models}
The authors of (\cite{bradley2018translating}) formalised the notion of a translation between functorial language models. We illustrate this construction with an example on translating simple noun phrases and the problems one may encounter. 
\begin{definition}
Let $(\mathcal{C}, \otimes, 1_\mathcal{C})$ and $(\mathcal{D}, \odot, 1_\mathcal{D})$ be monoidal categories. A \emph{monoidal functor} $F : \mathcal{C} \to \mathcal{D}$ is a functor equipped with a natural isomorphism $\Phi_{x,y} : F(x) \odot F(y) \to F(x \otimes y)$ for every pair of objects $x,y \in \mathcal{C}$ and an isomorphism $\phi: 1_{\mathcal{D}} \to F(1_\mathcal{C})$ such that for any triple of objects $x,y,z \in \mathcal{C}$, the following diagram commutes 
\begin{center}
\begin{tikzcd}[sep=small]
(F(x)\odot F(y))\odot F(z) \arrow[rr, "{\Phi_{x,y} \odot 1_{F(z)}}"] \arrow[d] &  & F(x\otimes y)\odot F(z) \arrow[rr, "{\Phi_{x\otimes y, z}}"]  &  & F((x\otimes y)\otimes z) \arrow[d] \\
F(x)\odot (F(y)\odot F(z)) \arrow[rr, "{1_{F(x)} \odot \Phi_{y,z}}"]           &  & F(x)\odot F(y \otimes z) \arrow[rr, "{\Phi_{x,y \otimes z}}"] &  & F(x\otimes (y \otimes z))         
\end{tikzcd}
\end{center}
where the vertical arrows apply the associativity in their respective categories. Moreover, for every object $x \in \mathcal{C}$, the following two squares commute:
\begin{center}
 \begin{tikzcd}[sep=small]
1_\mathcal{D} \odot F(x) \arrow[rr] \arrow[d] &  & F(x)                                &  & F(x) \odot 1_\mathcal{D} \arrow[rr] \arrow[d] &  & F(x)                                  \\
F(1_\mathcal{C}) \odot  F(x) \arrow[rr]       &  & F(1_\mathcal{C}\otimes x) \arrow[u] &  & F(x) \odot F(1_\mathcal{C}) \arrow[rr]        &  & F(x\otimes 1_{\mathcal{C}}). \arrow[u]
\end{tikzcd}   
\end{center}
\end{definition}
\begin{definition}
    Let $(F, \Phi, \phi)$ and $(G, \Psi, \psi)$ be monoidal functors between the monoidal categories $\mathcal{C}$ and $\mathcal{D}$. A \emph{monoidal natural transformation} $\alpha: F \Rightarrow G$ is a natural transformation where the following diagrams commute:
    \begin{center}
\begin{tikzcd}
F(x) \odot F(y) \arrow[rr, "\alpha(x) \odot \alpha(y)"] \arrow[d, "{\Phi_{x,y}}"] &  & G(x) \odot G(y) \arrow[d, "{\Psi_{x,y}}"] &  &   1_\mathcal{D} \arrow[d, "\phi"] \arrow[rd, "\psi"]  &                  \\
F(x\otimes y) \arrow[rr, "\alpha_{x \otimes y}"]                                  &  & G(x \otimes y)                            &  &   F(1_\mathcal{C}) \arrow[r, "\alpha(1_\mathcal{C})"] & G(1_\mathcal{C}).
\end{tikzcd}
    \end{center}
\end{definition}
\begin{definition}
Let $\mathcal{A} : P \to \fv$ and $\mathcal{B} : Q \to \fv$ be two functorial language models. A \emph{translation} from $F$ to $G$ is a tuple $(T, \alpha)$, where $T : P \to Q$ is a monoidal functor and $\alpha: \mathcal{A} \Rightarrow \mathcal{B} \circ T$ is a monoidal natural transformation. 
\end{definition}
\begin{eg}\label{ex:jpen adj noun}
We attempt to translate simple phrases of the type \textit{adjective + noun} from Japanese to English. We work on a restricted model. Let $J = \pgrp (\{s_J, n_J\})$ be the free pregroup (or  category) generated by the sentence and noun types in Japanese and let $E = \pgrp (\{s_E, n_E\})$ be the free pregroup generated by the sentence and noun types in English. 

The functorial language models are denoted by $\mathcal{J} : J \to \fv$ and $\mathcal{E} : E \to \fv$, respectively. The semantic assignment is straightforward: $\mathcal{J}(n_J) = N_J, \mathcal{J}(a_J) = A_J, \mathcal{E}(n_E) = N_E, \mathcal{E}(a_E) = A_E$. 

The translation will consist of the monoidal functor \[T : J \to E,\] which sends $s_J \mapsto s_E$ and $n_J \mapsto n_E$. Automatically, we have that the type reduction is preserved in the corresponding languages, i.e. $T\left((n_J n_J^\ell) n_J\right) = T(n_J) = n_E$. Due to monoidality, it suffices to define the components $\alpha_{n_J}, \alpha_{s_J}$ of the natural transformation $\alpha: \mathcal{J} \Rightarrow \mathcal{E} \circ T$ in order to parse semantics. 

Additionally, the natural transformation $\alpha$ must commute with the monoidal functor $T$. Pictorially we have a commutative square:
\begin{center}
\begin{tikzcd}
(N_J \otimes N_J) \otimes N_J \arrow[rr, "\mathcal{J}(\varepsilon^\ell_{N_J}\otimes 1_{\mathcal{J}})"] \arrow[d, "\alpha_{(n_J n_J^\ell) n_J}"] &  & N_J \arrow[d, "\alpha_{n_J}"] \\
(N_E \otimes N_E) \otimes N_E \arrow[rr, "\mathcal{E}(\varepsilon^\ell \otimes 1_\mathcal{E})"]                               &  & N_E .                         
\end{tikzcd}
\end{center}

Consider the concrete words $\vec{red} \in N_E \otimes N_E$, $\vec{cat} \in N_E$, $\vec{akai} \in N_J \otimes N_J$, and $\vec{neko} \in N_J$. The diagram says that if we first use Japanese grammar rules to reduce $\vec{akai}\otimes \vec{neko}$ to $\vec{akai\ neko}$ and then translate to $\vec{red\  cat}$ is the same thing as first translating component-wise $\vec{akai}\otimes \vec{neko}$ to $\vec{\vec{red}\otimes \vec{cat}}$ and then using English grammar rules to reduce to $\vec{red\ cat}$. 

Since there is no discrepancy in word order, this example of phrasal translation works in the desired way. If we instead wanted to translate the phrase ``\textit{akai neko}'' from Japanese into ``\textit{pisic\u{a} ro\c{s}ie}'' in Romanian, we would encounter some difficulties. The latter is a \textit{noun + adjective} phrase, as the natural word order in Romanian for such phrases is the opposite to the word order in Japanese.

The reduction rule in Romanian is given by \[n_R (n_R^r n_R) \to n_R.\] 

Suppose there exists a monoidal functor $T' : J \to R$ that takes Japanese grammar types to Romanian grammar types. Then we want to preserve the reduction rules, i.e. \[T'\left((n_J n_J^\ell) n_J\right) = n_R (n_R^r n_R).\] 

We thus obtain the condition: $T'\left(n_J^\ell\right) = n_R^r$. However, left and right adjoints must be preserved by a strong monoidal functor. Hence this condition cannot be fulfilled.

Section \ref{sec:translation} introduces techniques that can help us overcome such problems with word order changes. 

\end{eg}
\section{Japanese crash course}\label{sec:japanese}
\subsection{Generalities} Japanese is a synthetic and agglutinative language. The usual word order is subject-object-verb (SOV) with topic-comment sentence structure. There are no definite/indefinite articles and nouns do not possess either grammatical gender or number. Verbs and adjectives are conjugated for tense, voice, and aspect, but not person or number. Particles are attached to words to identify their grammatical role. We write sentences natively and employ the \emph{Nihon-siki} romanisation system.

The sentence ``\textit{The cat eats fish}'' can be represented in two different but closely related ways.

\begin{examples}
\item \trigloss{\zh{猫} \zh{が} \zh{魚} \zh{を} \zh{食べる}}{neko ga sakana wo taberu}{cat \textsc{NOM} fish \textsc{ACC} eat}{}

\item \trigloss{\zh{猫} \zh{は} \zh{魚} \zh{を} \zh{食べる}}{neko ha sakana wo taberu}{cat \textsc{TOP} fish \textsc{ACC} eat}{}
\end{examples}

Note the use of the subject particle ``ga,'' the topic particle ``ha,'' and the direct object particle ``wo.'' Remark that Japanese distinguishes between topic and subject. The topic generally needs to be explicitly introduced at the beginning of a discourse, but as the discourse carries on, the topic need not be the grammatical subject of every sentence. Both sentences translate into English as ``The cat eats fish,'' or ``Cats eat fish.'' However, a more pertinent interpretation of the second sentence is ``As for the cat/Speaking of the cat, it eats fish.'' 

Another important aspect of word order in Japanese is head finality. Phrases can be broadly described as consisting of a \textbf{head} and a \emph{modifier}. English is generally a head initial language. Consider for example the phrases: ``\textbf{to} school,'' ``\textbf{in} England,'' and ``red \textbf{cat}.'' The word that gets modified tends to come before the modifiers, the main exception being that nouns succeed the adjectives that modify them. In contrast, Japanese is per excellence a head final language. Our example phrases become

\trigloss[ex]{\zh{学校} \textbf{\zh{へ}}}{gakk\={o} \textbf{he}}{school \textbf{to}}{}

\trigloss[ex]{\zh{イギリス} \textbf{\zh{に}}}{igirisu \textbf{ni}}{England \textbf{in}}{}

\trigloss[ex]{\zh{赤い} \textbf{\zh{猫}}}{akai \textbf{neko}}{red \textbf{cat}}{}

Head finality is also encountered in the case of relative clauses, which usually occur before the part of speech they modify. This phenomenon is demonstrated by the following pair of phrases.
\begin{examples}
\item \trigloss{\zh{女} \zh{が} \zh{赤い} \zh{ワンピース} \zh{を} \zh{着た}}{onna ga akai wanp\^{i}su wo kita}{woman \textsc{NOM} red dress \textsc{ACC} wore}{The woman wore a red  dress}
\item \trigloss{\zh{赤い} \zh{ワンピース} \zh{を} \zh{着た} \textbf{\zh{女}} }{akai wanp\^{i}su wo kita \textbf{onna}}{red dress \textsc{ACC} wore \textbf{woman}}{\textbf{The woman}, who wore a red dress}
\end{examples}

This is a prime example of a structure where the word order is changed during translation. The following section will develop the algebraic machinery to interpret such translations.

Subjects are habitually dropped when they are clear from context, and personal pronouns are used sparingly. We conclude this section with an example, which demonstrates how a very common reflexive/personal pronoun ``zibun'' (``oneself'') can lead to ambiguous interpretations. ``Zibun'' is often used as a way for the speaker to refer either to themselves or to their interlocutor. The sentence

\begin{example}
\trigloss{\zh{自分} \zh{が} \zh{嘘つき} \zh{か}}{zibun ga usotuki ka}{oneself \textsc{NOM} liar \textsc{QUESTION}}{}
\end{example}
can be translated as either ``\textit{Am I a liar?}'' or ``\textit{Are you a liar?}'' in the absence of further context.

\subsection{Compositional model}
Define $J = \pgrp \left(\{\pi, n, s_1, s_2, s, o_1, \ldots \}\right)$ to be the pregroup of grammar types associated to Japanese. Following (\cite{cardinal2002algebraic}) with slight modifications, we define the following atomic types:
\begin{itemize}
    \item[-] $\pi$ pronoun,
    \item[-] $n$ noun,
    \item[-] $s_1, s_2$ imperfective/ perfective sentence,
    \item[-] $\overline{s}$ topicalised sentence,
    \item[-] $s$ sentence,
    \item[-] $o_1$ nominative case,
    \item[-] $o_2$ accusative case,
    \item[-] $o_3$ dative case,
    \item[-] $o_4$ genitive case,
    \item[-] $o_5$ locative case,
    \item[-] $o_6$ lative case,
    \item[-] $o_7$ ablative case,
    \item[-] etc.
\end{itemize}

We also impose the following reductions in $J$:
\begin{align*}
    & s_i \to s && \overline{s} \to s &&& n \to \pi.
\end{align*}

We now discuss how to assign types to various parts of speech. Revisiting the example sentence ``neko ga sakana wo taberu'' (``the cat eats fish''), the words ``neko'' and ``sakana'' are both nouns and thus have type $n$. The subject particle ``ga'' has type $\pi^r o_1$, the direct object particle ``wo'' has type $n^r o_2$ and the transitive verb ``taberu'' then has type $o_2^r o_1^r s_1$. The sentence then has type $n (\pi^r o_1) n (n^ro_2) (o_2^r o_1^r s_1)$ and we can derive the following type reductions:
\begin{align*}
    n (\pi^r o_1) n (n^r o_2) (o_2^r o_1^r s_1) &\to (n \pi^r) o_1 (\pi \pi^r) o_2 (o_2^r o_1^r s_1)\\
    &\to (\pi \pi^r) o_1 (\pi \pi^r) o_2 (o_2^r o_1^r s_1)\\
    &\to  o_1 o_2 (o_2^r o_1^r s_1)\\
    &\to o_1 (o_2 o_2^r) o_1^r s_1\\
    &\to o_1 o_1^r s_1\\
    &\to s_1\\
    &\to s
\end{align*}
to see that the sentence is well-formed and reduces to the correct grammatical type. Here we used the reductions $n \to \pi$ and $s_1 \to s$ together with different applications of the contraction morphism $\varepsilon$. Graphically, this type reduction can be seen in the following diagram, where a lower bracket indicates that a contraction morphism of the type $\varepsilon$ was applied. 
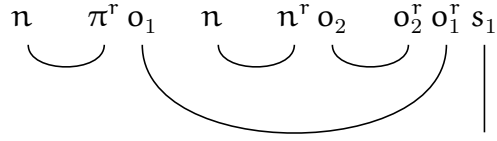
\begin{center}
\begin{tikzpicture}[semithick]
\tikzstyle{every state}=[
draw = black,fill = white, minimum size = 1mm]
\node (1) at (0,0) {$\fss{n}$};
\node (2) at (1,0) {$\fsub{\pi^r}$};
\node (3) at (1.5,0) {$\fsup{o_1}$};
\node (4) at (2.5,0) {$\fss{n}$};
\node (5) at (3.5,0) {$\fsub{n^r}$};
\node (6) at (4,0) {$\fsup{o_2}$};
\node (7) at (5,0) {$o_2^r$};
\node (8) at (5.5,0) {$o_1^r$};
\node (9) at (6,0) {$\fsup{s_1}$};
\node (10) at (6,-1.65) {};
\path[-,bend right=90,looseness=1] (1) edge node { } (2);
\path[-,bend right=90,looseness=1] (3) edge node { } (8);
\path[-,bend right=90,looseness=1] (4) edge node { } (5);
\path[-,bend right=90,looseness=1] (6) edge node { } (7);
\path[-] (9) edge node { } (10);
\end{tikzpicture}
\end{center}

Since word order is flexible, the same sentence could have been written as ``sakana wo neko ga taberu,'' and then \textit{taberu} would have been assigned the type $o_1^r o_2^r s_1$. As we want to take advantage of the \emph{Switching Lemma} while performing computations, we want to restrict ourselves to working with freely generated pregroups. Situations where certain words or verbs can be assigned different types are generally handled by adding \emph{metarules}. Informally, a metarule stipulates that if a grammar contains rules that match a specified pattern, then it also contains rules that match some other specified pattern. In our concrete example, we could impose the following metarule.

\begin{metarule}\label{metarule 1}
Any transitive verb that has type $o_1^ro_2^rs_i$ also has type $o_2^ro_1^rs_i$. 
\end{metarule}

Moving away from transitive verbs, the ablative particle ``kara'' has type $\pi^r o_7$ and the lative particle ``he'' has type $\pi^r o_6$. In the following example, the verb ``untensita'' has type $o_6^r o_7^r s_2$. 
\trigloss[ex]{\zh{家} \zh{から} \zh{駅} \zh{へ} \zh{運転した}}{ie kara eki he untensita}{house \textsc{ABL} station \textsc{LAT} drove}{(I) drove from home to the train station.}

Causative passive verbs take a subject and an indirect object marked with the dative particle ``ni'' of type $\pi^r o_3$. For instance, the verb ``yomaseta'' (``x made y read'') has type $o_2^ro_3^ro_1^rs_2$.

\trigloss[ex]{\zh{先生} \zh{が} \zh{私} \zh{に} \zh{本} \zh{を} \zh{読ませた}}{sensei ga watasi ni hon wo yomaseta}{teacher \textsc{NOM} I \textsc{DAT} book \textsc{ACC} read-\textsc{CAUSE-PAS}}{The teacher made me read the book.}

The genitive particle ``no'' has type $\pi^r o_4$ together with a metarule that states that type $o_4$ is equivalent to type $nn^\ell$. The possessor is always on the left in a genitive construction. The topic particle ``ha'' is distinguished from the subject particle ``ga'' and has type $\pi^r \overline{s}s^\ell$, i.e. ``ha'' requires a topic on the left and a sentence about the topic on the right. 
\trigloss[ex]{\zh{私} \zh{の} \zh{車} \zh{は} \zh{箸} \zh{を} \zh{渡れない}}{watasi no kuruma ha hasi wo watarenai}{I \textsc{GEN} car \textsc{TOP} bridge \textsc{ACC} cross-\textsc{POT-NEG}}{I cannot cross the bridge with my car/About my car, it cannot cross the bridge.}

In the latter example, the type reduction goes as follows:
\begin{align*}
    &\pi  (\pi^ro_4) n (\pi^r\overline{s}s^\ell) n (\pi^ro_2) (o_2^rs_1) && \\
    \to &(\pi\pi^r) o_4 n (\pi^r\overline{s}s^\ell) (n\pi^r)(o_2o_2^r)s && \text{associativity}\\
    \to &(1)(nn^\ell)n(\pi^r\overline{s}s^\ell)(n \pi^r)(1)s &&\text{contractions + genitive metarule}\\
    \to & n(n^\ell n)(\pi^r\overline{s}s^\ell)(\pi\pi^r)s &&n\to \pi\\
    \to & (n\pi^r)\overline{s}(s^\ell s) &&\text{associativity}\\
    \to & \overline{s} &&\text{contractions}\\
    \to & s.
\end{align*}
\section{Translation and decorated pregroups}\label{sec:translation}
\subsection{Decorated pregroups} As Example \ref{ex:jpen adj noun} shows, our initial machinery is not suited to translating phrases between languages with different word orders. The morphism of pregroups (or monoidal functor) $T : P \to Q$ that transfers information from the source language to the target language happens to be too rigid. 
We decorate pregroups with additional structures so that we can have more control over the monoid's operation. To this end, we define anti-homomorphisms for the purpose of inverting word order and pregroups with braces and $\beta$-pregroups to get more refined control over associativity. 

\begin{definition}
An \emph{anti-homomorphism of monoids} is a map $\Phi : P \to Q$ such that for all elements $x,y \in P$ we have $\Phi(xy) = \Phi(y)\Phi(x)$. 
\end{definition}

\begin{definition}
Let $(P,\cdot)$ be a monoid. The opposite monoid $(P^{\text{op}}, *)$ is the monoid which has the same elements as $P$ and the operation for all $x,y \in P^{\text{op}}$ is given by $x*y = y \cdot x$. It is elementary to observe that $(P, \cdot) \simeq (P^{\text{op}}, *)$. 
\end{definition}

In light of this, an anti-homomorphism can be viewed as a morphism from the opposite monoid $\Phi: P^{\text{op}} \to Q$. Additionally, an anti-homomorphism of pregroups takes left adjoints to right adjoints and vice-versa. 

\begin{eg}
In Example \ref{ex:jpen adj noun} the problem of translating ``adjective + noun'' phrases from Japanese into Romanian can be solved by setting the translation functor to be an anti-homomorphism that sends $n_J \mapsto n_R$. Then the functor $T$ preserves the desired reductions
\[T((n_J n_J^\ell) n_J) = T(n_J) \left(T(n_J^\ell) T(n_J)\right) = n_R \left(n_R^r n_R\right) \to n_R.\] 
\end{eg}

Parsing longer phrases and full sentences adds new layers of complexity. For instance, in simple short phrases there often is exactly one way of performing type reductions in order to assess the syntactic type of a phrase. Associativity can introduce ambiguity while parsing phrases. The following example demonstrates this.
\begin{eg}
Consider the phrase ``old teachers and students.'' We assign type $n$ to ``teachers'' and ``students.'' We assign the type $n n^\ell$ to the adjective  ``old.'' The conjunction ``and'' in this phrase requires two inputs of noun type to produce a noun phrase and is thus assigned $n^r n n^\ell$. We can use the associativity of the monoid operation to perform two distinct type reductions. 

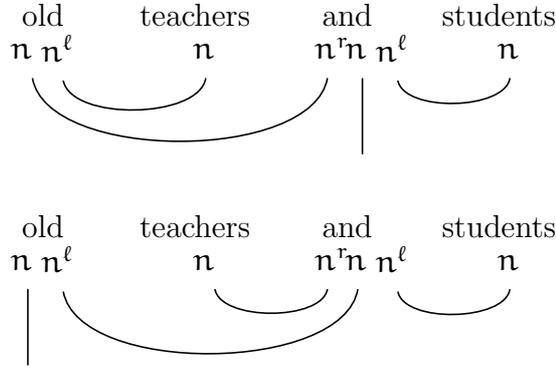
\begin{center}
\begin{tikzpicture}[semithick]
\tikzstyle{every state}=[
draw = black,fill = white]
\node (1) at (0,0) {old};
\node (2) at (2,0) {teachers};
\node (3) at (4,0) {and};
\node (4) at (6,0) {students};
\node (i) at (-0.2,-0.5) {$\fss{n}$};
\node (i') at (0.2,-0.5) {$\fsub{n^\ell}$};
\node (ii) at (2.2,-0.5) {$\fss{n}$};
\node (iii) at (3.8, -0.5) {$\fsub{n^r}$};
\node (iii') at (4.2,-0.5) {$\fss{n}$};
\node (iii'') at (4.6,-0.5) {$\fsub{n^\ell}$};
\node (iv) at (6.2, -0.5) {$\fss{n}$};
\node (f) at (4.2,-2) {};
\path[-,bend right=80,looseness=0.75] (i') edge node { } (ii);
\path[-,bend right=80,looseness=0.75] (i) edge node { } (iii);
\path[-,bend right=80,looseness=0.75] (iii'') edge node { } (iv);
\path[-] (iii') edge node { } (f);
\end{tikzpicture}
\end{center}
\begin{center}
\begin{tikzpicture}[semithick]
\tikzstyle{every state}=[
draw = black,fill = white]
\node (1) at (0,0) {old};
\node (2) at (2,0) {teachers};
\node (3) at (4,0) {and};
\node (4) at (6,0) {students};
\node (i) at (-0.2,-0.5) {$\fss{n}$};
\node (i') at (0.2,-0.5) {$\fsub{n^\ell}$};
\node (ii) at (2.2,-0.5) {$\fss{n}$};
\node (iii) at (3.8, -0.5) {$\fsub{n^r}$};
\node (iii') at (4.2,-0.5) {$\fss{n}$};
\node (iii'') at (4.6,-0.5) {$\fsub{n^\ell}$};
\node (iv) at (6.2, -0.5) {$\fss{n}$};
\node (f) at (-0.2,-2) {};
\path[-,bend right=80,looseness=0.75] (i') edge node { } (iii');
\path[-,bend right=80,looseness=0.75] (ii) edge node { } (iii);
\path[-,bend right=80,looseness=0.75] (iii'') edge node { } (iv);
\path[-] (i) edge node { } (f);
\end{tikzpicture}
\end{center}

Both type reductions give the desired noun phrase. However, the two interpretations are slightly different. The first one attributes the adjective ``old'' to ``teachers'' only, and so the sentence is parsed as ``(old teachers) and students,'' while the second type reduction attributes ``old'' to both ``teachers'' and ``students,'' giving the phrase ``old (teachers and students).''

One can construct examples where changing the order of reductions can make the difference between reducing down to a well-formed sentence and reducing down to a phrase that cannot be grammatically accepted. For this reason, one can add a modality or a $\beta$-structure to the pregroup to locally suppress associativity. This is to ensure that our phrases reduce to the correct type or that we distribute modifiers in a prescribed way. 
\end{eg}
Pregroups with modalities were first introduced in \cite{fadda2002towards} and their logic was more extensively studied in \cite{kislak2007logic}. 
\begin{definition}
A $\beta$-\emph{pregroup} is a pregroup $(P, \cdot, 1, -^\ell, -^r, \leq)$ together with a monotone mapping $\beta: P \to P$ such that $\beta$ has a right adjoint $\hat{\beta} : P \to P$, i.e. for all $x,y \in P$ we have $\beta(x) \leq y$ if and only if $x \leq \hat{\beta}(y)$. 
\end{definition}

In practice, we enrich our pregroup grammars with types with modalities to indicate certain reductions must be performed first. 

\begin{eg}
In our previous example, we can prescribe the parsing ``(old teachers) and students'' by assigning the types
\[n \left[\mbox{\boldmath$\beta$}(n)\right]^\ell \cdot \left[\mbox{\boldmath$\beta$}(n)\right] \cdot n^r n n^\ell \cdot n\]
and the parsing ``old (teachers and students)'' by assigning the types
\[n n^\ell \cdot \left[\mbox{\boldmath$\beta$}(n)\right] \cdot \left[\mbox{\boldmath$\beta$}(n)\right]^r n n^\ell \cdot n.\]
\end{eg}

We now have ways to invert word order globally and block associativity locally. We conclude this section by introducing a new type of decoration which allows us to locally control word order.

The reader is also encouraged to consult \cite{stabler2008tupled} for an introduction to tupled pregroups, \cite{lambek2010exploring} for an analysis of French sentences using products of pregroups, and \cite{vbo2023} for pregroups with local precyclicity. 

Next, we introduce a new pregroup decoration. 

\begin{definition}
A \emph{monoid with $k$-braces} $(P, \cdot, 1)$ is a free monoid in which every word is a prescribed concatenation of $k>0$ distinguished subwords. Extending this and subsequent definitions to \emph{pregroups with $k$-braces} is immediate.
\end{definition}
\begin{eg}
Consider the free monoid on two letters $F = \mathrm{Mon}(\{a,b\})$. Viewing $F$ as a monoid with $2$-braces, $\langle abba \rangle \langle b\rangle$ and $\langle abb \rangle \langle a b\rangle$ are distinct words because they have distinct distinguished subwords. 
\end{eg}
\begin{definition}
A \emph{morphism of monoids with $k$-braces} $f : (P, \cdot) \to (Q, *)$ is a morphism of monoids $f : (P, \cdot) \to (Q, *)$ which preserves distinguished subwords. In symbols: 
\[f\left(\langle w_1 \rangle \cdot \ldots \cdot \langle w_k \rangle\right) = \langle f(w_1) \rangle * \langle f(w_2) \rangle *\ldots *\langle f(w_k) \rangle .\]
\end{definition}
\begin{eg}[Some useful constructions] \label{eg:useful} We define two morphisms of monoids with braces which are useful in understanding translations. 

First, let $P$ be a monoid with $2$-braces and consider a word \[w = \langle w_1 \rangle \langle w_2 \rangle.\]
Since the underlying monoid of $P$ is free, we can view $w$ as an element of the free product $P * P \simeq P$ where the distinguished subword $w_i$ belongs to the $i$-th factor. Take the following sequence of monoid morphisms
{\small
\begin{center}
\begin{tikzcd}
             & \Psi:\ P \simeq P * P \arrow[r,"f"] & P\times P \arrow[r,"g"] & P^{\text{op}}\times P^{\text{op}} \arrow[r,"h"] & {} \\
{} \arrow[r,"h"] & P^{\text{op}}*P^{\text{op}} \arrow[r,"i"]       & Q     & {}                               
\end{tikzcd}                                                             
\end{center}
}
Here,  $f$ is the canonical surjection sending $\langle w_1\rangle\langle w_2\rangle \mapsto (w_1, w_2)$, $g$ is a pair of anti-isomorphisms which act like the identity on atomic types $(w_1, w_2) \mapsto (w_1^{\text{op}}, w_2^{\text{op}})$, $h$ is the canonical injection sending $(w_1^{\text{op}}, w_2^{\text{op}}) \mapsto \langle w_1^{\text{op}}\rangle \langle w_2^{\text{op}}\rangle$ and $i$ is some fixed homomorphism of monoids with $2$-braces. 

Secondly, let $P$ be a monoid with $3$-braces. We construct in a similar fashion the following morphism. 
{\small
\begin{center}
\begin{tikzcd}
             & \Xi:\ P \simeq P * P *P \arrow[r] & P\times P \times P \arrow[r] & P\times P^{\text{op}}\times P \arrow[r] & {} \\
{} \arrow[r] & P*P^{\text{op}}*P \arrow[r]       & P*P*P\simeq P \arrow[r]      & Q                                       &   
\end{tikzcd}
\end{center}
}
\end{eg}

We now proceed with concrete examples of phrasal translations. Throughout the remainder of the section, we work with two functorial language models: $\mathcal{J} : J \to \fv$ for Japanese and $\mathcal{E} : E \to \fv$ for English. We also impose the following useful metarule.

\begin{metarule}\label{metarule 3}
Any verb of type $s o_1^r w$ also has type $o_1^\ell s w$, where $w$ stands for all the remaining required complements. 
\end{metarule}

\subsection{``There is''/``There exists''}
Japanese has two verbs of existence, ``iru'' and ``aru,'' which are used for animate and inanimate beings, respectively. They both roughly mean ``to be,'' although a more common English translation is ``there is/there exists.''

Consider the following sentence.
\begin{example}
\trigloss{\zh{森} \zh{に} \zh{猫} \zh{が} \zh{いる}}{mori ni neko ga iru}{forest \textsc{LOC} cat \textsc{NOM} be}{}
\end{example}

A human translator has numerous ways of approaching this sentence. A standard and mot-a-mot SVO translation is ``A cat is in the forest.`` An easy SVO upgrade would be ``A cat lives in the forest.'' Considering that this is a short story meant for children, one could even opt for ``In the forest lives a cat'' to induce a fairy tale type atmosphere to the text. In this article, we choose to translate this using a straightforward anti-homomorphism and thus we aim for ``There is a cat in the forest.''

We work with the following reduced models for grammar: \[J = \pgrp(\{n, o_1, o_5, s\}) \text{ and } E = \pgrp(\{n_E, o_{1E}, o_{5E}, s_E\}).\] The translation functor at the level of syntax is given by the anti-homomorphism $T : J \to E$ which sends $n\mapsto n_E, o_1\mapsto o_{1E}, o_5 \mapsto o_{5E}, s\mapsto s_E$. At the level of semantics we have $F(n) = F(o_1) = F(o_5) = N, F(s) = S$ and $G(n_E) = G(o_{1E}) = G(o_{5E}) = N_E, G(s_E) = S_E$. 

In $J$ we have the type reduction $r = n(n^r o_5) n (n^r o_1) (o_1^r o_5^r s) \leq s$. After applying the translation functor $T$ we get:
\begin{align*}
    T(n(n^r o_5) n (n^r o_1) (o_1^r o_5^r s)) &= T(s) T(o_5^r) T(o_1^r) T(o_1) T(n^r) T(n) T(o_5) T(n^r) T(n)\\
    &= (s_E o_{5E}^\ell o_{1E}^\ell) o_{1E} (n_E^\ell n_E) o_{5E} (n_E^\ell n_E)\\
    &\to s_E o_{5E}^\ell (o_{1E}^\ell o_{1E}) o_{5E}\\
    &\to s_E (o_{5E}^\ell o_{5E})\\
    &\to s_E. 
\end{align*}

At the level of semantics we define the natural transformation $\alpha : \mathcal{J} \Rightarrow \mathcal{E}\circ T$ to act in the expected way, i.e. the map $N \to N_E$ sends $\vec{neko} \mapsto \vec{cat}$, $\vec{mori} \mapsto \vec{forest}$ and the map $S \to S_E$ sends $\vec{iru} \mapsto \vec{there\ is}$. We also impose $\vec{ga} \mapsto \vec{a}$ and  $\vec{ni} \mapsto \vec{in\ the}$. The commutativity of the following diagram is immediate. 
{\small
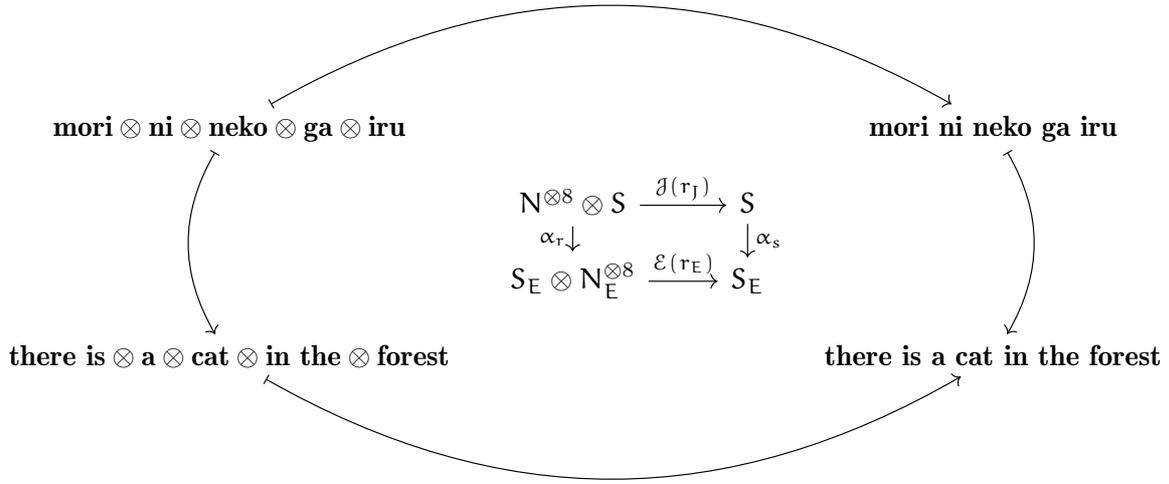
\begin{center}
\begin{tikzcd}[sep=small]
\vec{mori} \otimes \vec{ni} \otimes \vec{neko}\otimes \vec{ga} \otimes \vec{iru} \arrow[ddd, maps to, bend right] \arrow[rrrr, maps to, bend left] &                                                                                              &  &                         & \vec{mori\ ni\ neko\ ga\ iru} \arrow[ddd, maps to, bend left] \\
                                                                                                                                                   & N^{\otimes 8} \otimes S \arrow[d, "\alpha_{r}"'] \arrow[rr, "\mathcal{J}(r_J)"] &  & S \arrow[d, "\alpha_s"] &                                                               \\
                                                                                                                                                   & S_E \otimes N_E^{\otimes 8} \arrow[rr, "\mathcal{E}(r_E)"]                                             &  & S_E                     &                                                               \\
\vec{there\ is}\otimes \vec{a} \otimes \vec{cat} \otimes \vec{in\ the}\otimes \vec{forest} \arrow[rrrr, maps to, bend right]                       &                                                                                              &  &                         & \vec{there\ is\ a\ cat\ in\ the\ forest}                     
\end{tikzcd}
\end{center}
}


\subsection{Simple SOV sentences}
We describe a procedure for translating the following sentence.

\begin{example}
\trigloss{\zh{医者} \zh{は} \zh{手紙} \zh{を} \zh{書く}}{issya ga tegami wo kaku}{doctor \textsc{NOM} letter \textsc{ACC} write}{The doctor writes a letter.}
\end{example}

We work with the grammars \[J = \pgrp(\{n, o_1, o_2, s\}) \text{ and } E = \pgrp(\{n_E, o_{1E}, o_{2E}, s_E\}).\] The words ``issya'' and ``tegami'' are assigned the noun type $n$, the particles ``ga'' and ``wo'' have the usual types $n^r o_1$ and $n^r o_2$, respectively, and the transitive verb ``kaku'' has type $o_2^r o_1^r s$. The sentence is clearly well-formed: $n(n^r o_1) n (n^r o_2) (o_2^r o_1^r s) \to s$. 

Here we employ the notion of a pregroup with $2$-braces. In principle, for an SOV sentence we assign braces as follows: $\langle S \rangle \langle OV \rangle$. In our particular sentence, this becomes \[\big \langle n (n^r o_1) \big \rangle \big \langle n (n^r o_2) (o_2^r o_1^r s) \big \rangle.\]

We define our translation functor $\Psi$ to be the morphism of monoids with braces defined in Example \ref{eg:useful}. Together with Metarule \ref{metarule 3} this gives:

\begin{align*}
    \Psi \big \langle n (n^r o_1) \big \rangle \big \langle n (n^r o_2) (o_2^r o_1^r s) \big \rangle &= \big \langle (o_{1E} n_E^\ell) n_E \big \rangle \big \langle (s_E o_{1E}^\ell o_{2E}^\ell) (o_{2E} n_E^\ell) n \big \rangle \\
    &= \big \langle (o_{1E} n_E^\ell) n_E \big \rangle \big \langle (o_{1E}^r s_E o_{2E}^\ell) (o_{2E} n_E^\ell) n \big \rangle
\end{align*}

Then $\alpha$ can be defined on atomic types as follows: $\vec{issya} \mapsto \vec{doctor}$, $\vec{tegami}\mapsto \vec{letter}$, $\vec{kaku} \mapsto \vec{write}$, and the translation $(\Psi, \alpha)$ gives 
\[``\text{(A/The) doctor write(s) (a/the) letter.}''\]

Again, the articles and the conjugation of ``write'' into third person singular can either be added by brute force in our model by adding meanings to the particles ``ga'' and ``wo,'' or one can verify agreement and articles separately as a different step in the translation process. 

\subsection{Relative clauses} Interpreting relative pronouns in various languages in terms of pregroups proves to be quite challenging. In \cite{sadrzadeh2013frobenius} and \cite{sadrzadeh2014frobenius}, the authors add the additional structure of a Frobenius algebra on the pregroup. Informally, a Frobenius algebra structure enriches the $\varepsilon, \eta$ functorial yoga with additional maps, the most important of which are called ``copying map'' and ``uncopying map.'' These new morphisms allow one to better keep track of information inside a phrase. For instance, in the English sentence
\[``\text{The woman, who drove from Tokyo today, was late to the party}''\]
the new morphisms can formalise the fact that the subject of the main clause ``The woman was late to the party'' and the subject of the relative clause ``who drove from Tokyo today'' are one and the same. The relative pronoun ``who'' acts as a bridge that ``copies'' the subject into the relative clause and then transfers it back into the main clause. 

We translate the following relative clause. 

\begin{example}
\trigloss{\zh{今日} \zh{東京} \zh{から}  \zh{運転した} \zh{女}}{ky\={o} t\={o}ky\={o} kara untensita onna}{today Tokyo \textsc{ABL} drove woman}{The woman who drove from Tokyo today.}
\end{example}

We assign types in a less straightforward way. We first insert an empty word between the modifier ``t\={o}ky\={o} kara untensita'' and the head ``onna.'' We assign the following types: ``t\={o}ky\={o}'' and ``onna'' are both type $n$, the ablative particle ``kara'' has type $n^r o_7$, ``ky\={o}'' has type $t$ (temporal adverb), the verb ``untensita'' has type $o_7^rt^rso_1^\ell$ and the empty word acts like a phantom relative pronoun with type $o_1 s^r n n^\ell$. 

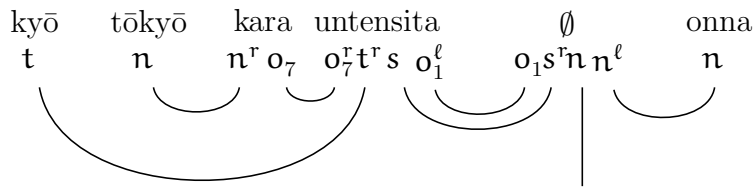
\begin{center}
\begin{tikzpicture}[semithick]
\tikzstyle{every state}=[
draw = black,fill = white]
\node (1) at (0,0) {ky\={o}};
\node (2) at (1.5,0) {t\={o}ky\={o}};
\node (3) at (3,0.05) {kara};
\node (4) at (4.5,0.05) {untensita};
\node (5) at (7,0) {$\emptyset$};
\node (6) at (9,0) {onna};
\node (7) at (0,-0.5) {$\fss{t}$};
\node (8) at (1.5,-0.5) {$\fss{n}$};
\node (9) at (2.75,-0.5) {$\fsub{n^r}$};
\node (9') at (3.25, -0.5) {$\fsup{o_7}$};
\node (10) at (4,-0.5) {$o_7^r$};
\node (10') at (4.4,-0.5) {$\fsub{t^r}$};
\node (10'') at (4.8, -0.5) {$\fss{s}$};
\node (10''') at (5.2, -0.5) {$o_1^\ell$};
\node (11) at (6.5,-0.5) {$\fsup{o_1}$};
\node (11') at (6.85,-0.5) {$\fsub{s^r}$};
\node (11'') at (7.2,-0.5) {$\fss{n}$};
\node (11''') at (7.55,-0.5) {$\fsub{n^\ell}$};
\node (12) at (9,-0.5) {$\fss{n}$};
\node (13) at (7.2, -2.3) {};
\path[-,bend right=80,looseness=1] (8) edge node { } (9);
\path[-,bend right=80,looseness=1] (7) edge node { } (10');
\path[-,bend right=80,looseness=1] (9') edge node { } (10);
\path[-,bend right=80,looseness=1] (10''') edge node { } (11);
\path[-,bend right=80,looseness=1] (10'') edge node { } (11');
\path[-,bend right=80,looseness=1] (11''') edge node { } (12);
\path[-] (11'') edge node { } (13);
\end{tikzpicture}
\end{center}

This construction generates a noun phrase and it can be translated using a straightforward anti-homomorphism. The advantage of this underhanded construction is that now we can translate the empty word as the relative pronoun ``who'' or ``that.'' This ties in perfectly with the Frobenius algebra approach of \cite{sadrzadeh2013frobenius}. In this example we modelled our relative clause as what the authors of the reference call a subject relative clause.
\subsection{Coordinate sentences} The simplest way of coordinating sentences is by connecting them with the particle ``ga'' (``and'') to which we assign the type $s^r s s^\ell$. We translate the following sentence where subjects are omitted. 

\begin{example}
\trigloss{\zh{家} \zh{に} \zh{着いた} \zh{が} \zh{手紙} \zh{を} \zh{書いた}}{ie ni tuita ga tegami wo kaita}{house \textsc{LOC} arrived and letter \textsc{ACC} wrote}{I arrived home and wrote a letter.}
\end{example}

In Japanese we have the following reduction diagram.

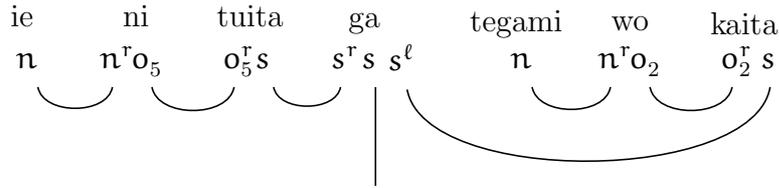
\begin{center}
\begin{tikzpicture}[semithick]
\tikzstyle{every state}=[
draw = black,fill = white]
\node (1) at (0,0.1) {ie};
\node (2) at (1.5,0.1) {ni};
\node (3) at (3,0.1) {tuita};
\node (4) at (4.5,0) {ga};
\node (5) at (6.5,0) {tegami};
\node (6) at (8,0) {wo};
\node (7) at (9.5,0) {kaita};
\node (i) at (0.15,-0.5) {$\fss{n}$};
\node (ii) at (1.25,-0.5) {$\fsub{n^r}$};
\node (ii') at (1.65,-0.5) {$\fsup{o_5}$};
\node (iii) at (2.85, -0.5) {$o_5^r$};
\node (iii') at (3.25,-0.5) {$\fss{s}$};
\node (iv) at (4.25,-0.5) {$\fsub{s^r}$};
\node (iv') at (4.65, -0.5) {$\fss{s}$};
\node (iv'') at (5, -0.5) {$\fsub{s^\ell}$};
\node (v) at (6.65,-0.5) {$\fss{n}$};
\node (vi) at (7.8,-0.5) {$\fsub{n^r}$};
\node (vi') at (8.2,-0.5) {$\fsup{o_2}$};
\node (vii) at (9.4,-0.5) {$o_2^r$};
\node (vii') at (9.9,-0.5) {$\fss{s}$};
\node (f) at (4.65, -2.3) {};
\path[-,bend right=80,looseness=1] (i) edge node { } (ii);
\path[-,bend right=80,looseness=1] (ii') edge node { } (iii);
\path[-,bend right=80,looseness=1] (iii') edge node { } (iv);
\path[-,bend right=80,looseness=0.7] (iv'') edge node { } (vii');
\path[-,bend right=80,looseness=1] (v) edge node { } (vi);
\path[-,bend right=80,looseness=1] (vi') edge node { } (vii);
\path[-] (iv') edge node { } (f);
\end{tikzpicture}
\end{center}

We decorate the pregroup with braces and assign the following type
\[\bigg\langle n \cdot n^r o_5 \cdot o_5^r s \bigg\rangle \bigg\langle s^r s s^\ell \bigg\rangle\bigg\langle n \cdot n^r o_2 \cdot o_2^r s \bigg\rangle.\]

Extending the morphism $\Psi$ from Example \ref{eg:useful} to monoids with $3$-braces, we obtain
\[\Psi \langle n \cdot n^r o_5 \cdot o_5^r s \rangle \langle s^r s s^\ell \rangle\langle n \cdot n^r o_2 \cdot o_2^r s \rangle = \langle s_E o_{5E}^\ell \cdot o_{5E}n_E^\ell \cdot n_E \rangle  \langle s_E^r s_E s_E^\ell \rangle \langle s_E o_{2E}^\ell \cdot o_{2E} n_E^\ell \cdot n_E \rangle.\]

Working under the assumption that an ommitted subject refers to the first person singular, the translation, after applying a suitably defined $\alpha$, is
\[``\text{(I) arrived home and (I) wrote (a) letter}.''\]
\subsection{Putting it all together}
We combine all our techniques to study a more complex sentence.

\begin{example}
\trigloss{\zh{制服} \zh{を} \zh{着た} \zh{学生} \zh{が} \zh{机} \zh{に} \zh{あった} \zh{本} \zh{を} \zh{盗んだ}}{seihuku o kita gakusei ga tukue ni atta hon wo nusunda}{uniform \textsc{ACC} wore student \textsc{NOM} desk \textsc{LOC} was book \textsc{ACC} stole}{The student, who wore a uniform, stole the book, which was on the desk.}
\end{example}

This is a standard SOV sentence, where both the subject and the direct object are modified by relative clauses. In the Japanese pregroup grammar, we have the following straightforward reductions. 
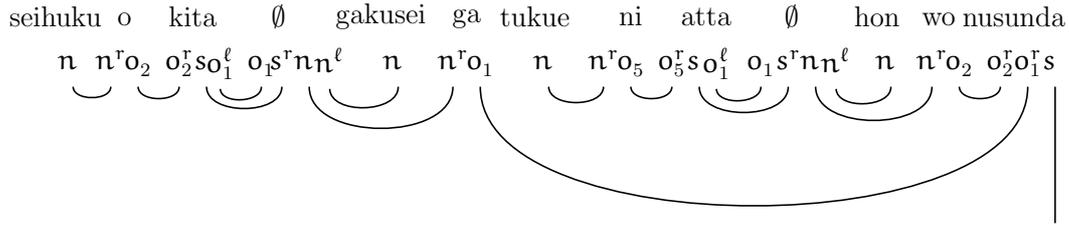
\begin{center}
\begin{tikzpicture}[semithick, scale=0.9, every node/.style={scale=0.9}]
\tikzstyle{every state}=[
draw = black,fill = white]
\node (1) at (0,0.2) {seihuku};
\node (2) at (1,0.15) {o};
\node (3) at (2,0.2) {kita};
\node (4) at (3.25,0.2) {$\emptyset$};
\node (5) at (4.75,0.2) {gakusei};
\node (6) at (6,0.15) {ga};
\node (7) at (7,0.2) {tukue};
\node (8) at (8.4,0.2) {ni};
\node (9) at (9.5,0.2) {atta};
\node (10) at (10.75,0.2) {$\emptyset$};
\node (11) at (12,0.2) {hon};
\node (12) at (12.9,0.15) {wo};
\node (13) at (14,0.2) {nusunda};
\node (i) at (0.25,-0.5) {$\fss{n}$};
\node (ii) at (0.8,-0.5) {$\fsub{n^r}$};
\node (ii') at (1.2,-0.5) {$\fsup{o_2}$};
\node (iii) at (1.8,-0.5) {$o_2^r$};
\node (iii') at (2.2,-0.5) {$\fss{s}$};
\node (iii'') at (2.4,-0.5) {$o_1^\ell$};
\node (iv) at (3,-0.5) {$\fsup{o_1}$};
\node (iv') at (3.3, -0.5) {$\fsub{s^r}$};
\node (iv'') at (3.7,-0.5) {$\fss{n}$};
\node (iv''') at (4,-0.5) {$\fsub{n^\ell}$};
\node (v) at (5,-0.5) {$\fss{n}$};
\node (vi) at (5.8, -0.5) {$\fsub{n^r}$};
\node (vi') at (6.2, -0.5) {$\fsup{o_1}$};
\node (vii) at (7.2,-0.5) {$\fss{n}$};
\node (viii) at (8,-0.5) {$\fsub{n^r}$};
\node (viii') at (8.4,-0.5) {$\fsup{o_5}$};
\node (ix) at (9,-0.5) {$o_5^r$};
\node (ix') at (9.4,-0.5) {$\fss{s}$};
\node (ix'') at (9.65,-0.5) {$o_1^\ell$};
\node (x) at (10.3,-0.5) {$\fsup{o_1}$};
\node (x') at (10.7,-0.5) {$\fsub{s^r}$};
\node (x'') at (11.1,-0.5) {$\fss{n}$};
\node (x''') at (11.4,-0.5) {$\fsub{n^\ell}$};
\node (xi) at (12.2,-0.5) {$\fss{n}$};
\node (xii) at (12.8,-0.5) {$\fsub{n^r}$};
\node (xii') at (13.2,-0.5) {$\fsup{o_2}$};
\node (xiii) at (13.8,-0.5) {$o_2^r$};
\node (xiii') at (14.2,-0.5) {$o_1^r$};
\node (xiii'') at (14.6,-0.5) {$\fss{s}$};
\node (f) at (14.6, -3) {};
\path[-,bend right=90,looseness=1] (i) edge node { } (ii);
\path[-,bend right=90,looseness=1] (ii') edge node { } (iii);
\path[-,bend right=90,looseness=1] (iii'') edge node { } (iv);
\path[-,bend right=90,looseness=1] (iii') edge node { } (iv');
\path[-,bend right=90,looseness=1] (iv''') edge node { } (v);

\path[-,bend right=90,looseness=1] (vii) edge node { } (viii);
\path[-,bend right=90,looseness=1] (viii') edge node { } (ix);
\path[-,bend right=90,looseness=1] (ix'') edge node { } (x);
\path[-,bend right=90,looseness=1] (ix') edge node { } (x');
\path[-,bend right=90,looseness=1] (x''') edge node { } (xi);

\path[-,bend right=90,looseness=1] (iv'') edge node { } (vi);
\path[-,bend right=90,looseness=1] (x'') edge node { } (xii);
\path[-,bend right=90,looseness=1] (xii') edge node { } (xiii);
\path[-,bend right=90,looseness=0.75] (vi') edge node { } (xiii');
\path[-] (xiii'') edge node { } (f);
\end{tikzpicture}
\end{center}

One may observe that in the diagram above we use associativity to our advantage to prove that the sentence reduces to the correct syntactic type. To get a failsafe reduction and translation we decorate our pregroup grammar with braces and a $\beta$-structure. The sentence is then assigned the type

\[\big \langle n \cdot n^r o_2 \cdot o_2^r o_1^\ell \cdot o_1 s^r n \mbox{\boldmath$\beta$}(n^\ell) \cdot  \mbox{\boldmath$\beta$}(n) \cdot n^r o_1\big \rangle \big\langle n \cdot n^r o_5 \cdot  o_5^r s o_1^\ell \cdot  o_1 s^r n \mbox{\boldmath$\beta$}(n^\ell) \cdot  \mbox{\boldmath$\beta$}(n) \cdot n^r o_2 \cdot o_2^ro_1^rs \big \rangle\]
and after applying the morphism $\Psi$ from Example \ref{eg:useful} together with Metarule \ref{metarule 3}, the sentence translates to
\begin{center}``(A/The) student, who wore (a/the) uniform, stole (a/the) book, which was (\textsc{LOC}) desk.''
\end{center}

\subsection{A Farsi to Japanese example} 
Farsi has certain similarities to Japanese which make translations (at the syntactic level, at least) somewhat simpler. For instance, Farsi also has SOV word order, nouns do not possess grammatical gender, and it is a pro-drop language. A key structural difference is that Farsi uses both prepositions and postpositons as case markers. 

Following \cite{sadrzadeh2007persian}, we use the following (reduced) pregroup to model Farsi grammar $F = \pgrp(\{\nu,\sigma,o,w\})$, where atomic types represent nouns, sentences, direct objects, and prepositional phrases respectively. On the Japanese side, we use $J = \pgrp(\{n,s,o_2,o_5\})$, with the usual meanings. Denote the two functorial language models as $\mathcal{F} : F \to \fv$ and $\mathcal{J} : J \to \fv$. 

We are interested in translating the following sentence from Farsi to Japanese.

\digloss[ex]{ket\={a}b r\={a} d\v{a}r b\={a}z\={a}r xarid}{book \textsc{ACC} \textsc{PREP} market bought}{He/She bought a book from the market.}

Here ``ket\={a}b r\={a}'' is the direct object, ``d\v{a}r b\={a}z\={a}r'' is the prepositional phrase and ``xarid'' is the transitive verb in the past tense. This example sentence drops the subject and uses both a postposition ``r\={a}'' and a preposition ``d\v{a}r'' to mark cases. In Farsi, we have the following reduction.
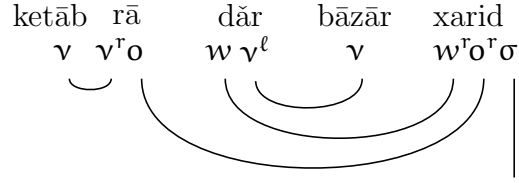
\begin{center}
    \begin{tikzpicture}[semithick]
\tikzstyle{every state}=[
draw = black,fill = white]
\node (1) at (0,0) {ket\={a}b};
\node (2) at (1,0) {r\={a}};
\node (3) at (2.5,0) {d\v{a}r};
\node (4) at (4,0) {b\={a}z\={a}r};
\node (5) at (5.5,0) {xarid};
\node (i) at (0.25,-0.5) {$\fss{\nu}$};
\node (ii) at (0.8,-0.5) {$\fsub{\nu^r}$};
\node (ii') at (1.2,-0.5) {$\fss{o}$};
\node (iii) at (2.3,-0.5) {$\fss{w}$};
\node (iii') at (2.7,-0.5) {$\fsub{\nu^\ell}$};
\node (iv) at (4.1,-0.5) {$\fss{\nu}$};
\node (v) at (5.3,-0.5) {$\fsub{w^r}$};
\node (v') at (5.7, -0.5) {$\fsub{o^r}$};
\node (v'') at (6.1,-0.5) {$\fss{\sigma}$};
\node (f) at (6.1, -2.3) {};
\path[-,bend right=90,looseness=0.9] (i) edge node { } (ii);
\path[-,bend right=90,looseness=0.9] (iii') edge node { } (iv);
\path[-,bend right=90,looseness=0.9] (iii) edge node { } (v);
\path[-,bend right=90,looseness=0.9] (ii') edge node { } (v');
\path[-] (v'') edge node { } (f);
\end{tikzpicture}
\end{center}

The functorial language models $\mathcal{F}, \mathcal{J}$ send $\nu, o, w \mapsto N_F$ (Farsi nouns) and $\sigma \mapsto S_F$ (Farsi sentences), and also $n, o_2, o_5 \mapsto N_J$ (Japanese nouns) and $s \mapsto S_J$ (Japanese sentences). The natural transformation $\alpha: \mathcal{F} \Rightarrow \mathcal{J} \circ i$ is given by $\vec{ket\overline{a}b} \mapsto \vec{hon}$, $\vec{r\overline{a}} \mapsto \vec{o}$, $\vec{d\breve{a}} \mapsto \vec{de}$, $\vec{b\overline{a}z\overline{a}r} \mapsto \vec{itiba}$ and $\vec{xarid} \mapsto \vec{kaimasita}$. At the syntactic level, we define some monoidal translation functor $T : F \to J$ which takes $\nu \mapsto n, \sigma \to s, o \to o_2,$ and $w \to o_5$. 

The pregroups are decorated with $3$-braces. The sentence is assigned the type
\[\big\langle \nu \cdot \nu^r o \big\rangle \big\langle w \nu^\ell \cdot \nu \big\rangle \big\langle w^ro^r \sigma\big\rangle.\]

Syntactically, the translation functor is taken to be $\Xi$ from Example \ref{eg:useful}. The word order is altered as follows
\[\Xi\big\langle \nu \cdot \nu^r o \big\rangle \big\langle w \nu^\ell \cdot \nu \big\rangle \big\langle w^ro^r \sigma\big\rangle = \big\langle n \cdot n^r o_2 \big\rangle \big\langle n \cdot n^r o_5 \big\rangle \big\langle o_5^ro_2^r s\big\rangle\]
which leads to the following type reduction in Japanese.
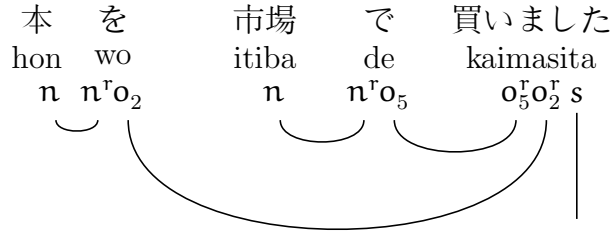
\begin{center}
        \begin{tikzpicture}[semithick]
\tikzstyle{every state}=[
draw = black,fill = white]
\node (J1) at (0,0.5) {\zh{本}};
\node (J2) at (1,0.5) {\zh{を}};
\node (J3) at (3,0.5) {\zh{市場}};
\node (J4) at (4.5,0.5) {\zh{で}};
\node (J5) at (6.5,0.5) {\zh{買いました}};
\node (1) at (0,0) {hon};
\node (2) at (1,0) {wo};
\node (3) at (3,0) {itiba};
\node (4) at (4.5,0) {de};
\node (5) at (6.5,0) {kaimasita};
\node (i) at (0.25,-0.5) {$\fss{n}$};
\node (ii) at (0.8,-0.5) {$\fsub{n^r}$};
\node (ii') at (1.2,-0.5) {$\fsup{o_2}$};
\node (iii) at (3.2,-0.5) {$\fss{n}$};
\node (iii') at (4.3,-0.5) {$\fsub{n^r}$};
\node (iv) at (4.7,-0.5) {$\fsup{o_5}$};
\node (v) at (6.3,-0.5) {$o_5^r$};
\node (v') at (6.7, -0.5) {$o_2^r$};
\node (v'') at (7.1,-0.5) {$s$};
\node (f) at (7.1, -2.3) {};
\path[-,bend right=90,looseness=0.9] (i) edge node { } (ii);
\path[-,bend right=90,looseness=0.9] (iii) edge node { } (iii');
\path[-,bend right=90,looseness=0.9] (iv) edge node { } (v);
\path[-,bend right=90,looseness=0.9] (ii') edge node { } (v');
\path[-] (v'') edge node { } (f);
\end{tikzpicture}
\end{center}

\section{Future work}
In this article, we introduced decorated pregroups and used them as a means of constructing a compositional notion of translation between natural languages with different word order. The aim was to demonstrate that one can maintain a categorical approach to modelling translation without compromising on functoriality altogether. Some of our constructions are ad-hoc and there is room for improving most of them. 

First, there is the issue of translating between a language where nouns do not have grammatical gender and number to a language that does. Using product pregroups or tupled pregroups to handle grammatical agreement could be a way forward, although a straightforward model for achieving this appears elusive. 

Secondly, one could study translations between languages which have more featural and structural differences. For example, how could we interpret (functorially) translations between a language which has nominative-accusative alignment and a language that has ergative-absolutive (or split-ergative) alignment? 

Thirdly, this article heavily focuses on syntax. It would be interesting to model how meaning in translation can be negotiated between different speakers and how one can keep track of their evolving semantic spaces. On a more technical note, one could change the meaning space from $\fv$ to a category that possesses more substantial structure such as $\mathrm{ConvexRel}$, the category where the objects are \emph{convex algebras} and the morphisms as \emph{convex relations}. In \cite{bolt2019interacting} the authors showed that $\mathrm{ConvexRel}$ is a compact closed symmetric monoidal category and is thus suitable for modelling semantics in a compositional distributional functorial language model.

Finally, separate from the question of translation, some attention could be dedicated to expanding the work of Cardinal (\cite{cardinal2002algebraic} \cite{cardinal2006type} \cite{cardinal2007pregroup}) and producing a more complete pregroup approach to analysing other aspects of grammar that are typical to Japanese. In particular, the structure of coordinate and subordinate sentences and internally headed relative clauses are of particular interest to the author. 
\bibliography{article}
\bibliographystyle{alpha}
\end{document}